# Words that Matter: The Impact of Negative Words on News Sentiment and Stock Market Index

Wonseong Kim [a,b]

[a]  Korea University, 13, Jongam-ro, Seongbuk-gu, Seoul, Republic of Korea

[b]  University of St. Gallen, Rosenbergstrasse 30, 9000 St.Gallen, Switzerland

*E-mail addresses:*

wonseongkim@korea.ac.kr *(W. Kim, affiliated in Korea University),*

*\* The research was conducted during Kim's tenure as a visiting research fellow at the University of St. Gallen.*

## Abstract

This study investigates the impact of negative words on sentiment analysis and its effect on the South Korean stock market index, KOSPI200. The research analyzes a dataset of 45,723 South Korean daily economic news articles using Word2Vec, cosine similarity, and an expanded lexicon. The findings suggest that incorporating negative words significantly increases sentiment scores' negativity in news titles, which can affect the stock market index. The study reveals that an augmented sentiment lexicon (Sent1000), including the top 1,000 negative words with high cosine similarity to 'Crisis,' more effectively captures the impact of news sentiment on the stock market index than the original sentiment lexicon (Sent0). The results underscore the importance of considering negative nuances and context when analyzing news content and its potential impact on market dynamics and public opinion.

1. **Introduction**

Natural Language Processing (NLP) has increasingly become a pivotal tool in social science including economic analysis, as it allows social scientists to work with unstructured text data and explore new research fields. One such area of interest is the nexus between news sentiment and financial markets, which has gained significant attention in recent years. A growing body of research has demonstrated the effects of news sentiments on various aspects of financial markets, which include stock prices and returns, trading volume, and market volatility (Tetlock, 2007; Engelberg & Parsons, 2011; Loughran & McDonald, 2011; Heston & Sinha, 2017; Allen et al., 2019).

The influence of negative words on investor sentiment and decision-making has also garnered increasing interest among researchers. (Lucey & Dowling, 2005; Fenton et al., 2011; Baker & Yi, 2016) Negative words can shape readers' perceptions and potentially affect their investment decisions, highlighting the importance of understanding the role of language in economic contexts. In this study, we focus on the impact of negative words, specifically *Words that Matter*, on sentiment analysis in a dataset of 45,723 daily economic news articles from South Korea. By examining the effects of emotionally charged words and their impact on sentiment scores, we aim to shed light on the role of negative words in shaping market dynamics and investor sentiment. Our findings contribute to the literature on news sentiment and financial markets by informing future research on the underlying mechanisms through which news sentiment influences market behavior and investor decision-making.

Research in the field of economics has demonstrated that news sentiment can serve as a powerful predictor of market movements (Tetlock, 2007; Garcia, 2013). Furthermore, studies have found that the media's tone and language can significantly influence investor sentiment and their subsequent investment decisions (Baker & Wurgler, 2007; Fang & Peress, 2009). In particular, negative news has been shown to have a stronger impact on market reactions compared to positive news (Soroka & McAdams, 2015). The primary objective of this study is to examine how negative words, particularly words with heightened negativity, influences news sentiment and its subsequent impact on the KOSPI200 index, the leading benchmark index for Korean stock markets. To achieve this, we employ Word2Vec, cosine similarity, and an expanded lexicon to analyze the effect of 'Words that Matter' on the sentiment scores of news titles. Our findings reveal that incorporating negative words into the lexicon significantly amplifies sentiment scores' intensity, particularly the negativity of news titles.

    Section 2 provides a literature review on sentiment analysis in financial markets and the role of negative words in shaping investor sentiment. Section 3 outlines the methodology used to construct the sentiment lexicon and analyze the dataset of news articles. Section 4 presents our findings and discusses the implications of negative words on stock market indices, specifically the KOSPI200. Finally, Section 5 concludes with agenda for future research.

**Sentiment Index with Bag of Negative Words**

In order to examine the impact of negative words, we collect a dataset of 45,723 daily economic news articles from South Korea, spanning from January 4th, 1990 to December 31st, 2021. The Korea Press Foundation's BIG KINDS platform (www.bigkinds.or.kr) serves as the original data source, allowing for the extraction of news articles using specified keywords, media types, and geographic classifications. Our analysis focuses on online daily news from 54 major media outlets over the course of 31 years. The news data is processed using the Word2Vec language model (Mikolov et al., 2013). Developed by Google, this natural language processing algorithm aims to learn word embeddings[1] from large text datasets, such that words appearing in similar contexts are represented by "similar" vector values. Investigating word embeddings in a vector space can provide valuable insights into potential instances of negative words and problematic usage patterns. To identify negative words associated with specific keywords, cosine similarity is employed to measure the contextual similarity between two vectors (embedded words). Cosine similarity is computed as the cosine of the angle between two vectors in a multi-dimensional space, ranging from -1 to 1, with values closer to 1 indicating a higher degree of similarity.

    We now elaborate the data construction process. The initial step involves pre-processing of unstructured data, specifically the news titles. We propose extracting news sentiment from their titles, as they are likely the most representative feature of the overall news contents. A total of 45,723 news titles were cleaned using regular expressions (RegEx)

---

[1] Word Embedding is the process of transforming word characters to real numbers, and phrases in vector dimensions. The numbers capture their meaning and context in the sentence. Word2Vec use neural network structure to calculate word embeddings form input data (news titles). The model's goal is to learn the parameters that maximize the likelihood of predicting context words given a target word (Skip-Gram) or predicting a target word given context words (Continuous Bag-of-Words).

and *stop-words*, which is subsequently tokenized[2] using Open Korean Text (OKT[3]). OKT is an open-source Korean NLP library developed by the Korean company, SK Telecom. Key feature of OKT is that provides the Korean language's unique grammatical features. This process results in 3,795 tokenized words. The average sentence length of news titles is 9.8 words, with the longest one containing 64 words. The third step involves vectorizing the tokenized words into a multi-dimensional space using the Word2Vec model.[4] In particular, we employ the skip-gram method, known for its effectiveness in identifying semantic similarities between words (see Appendix A). This approach utilizes a neural network with a single hidden layer for embedding. The input layer consists of target words encoded as one-hot vectors, while the output layer comprises the context words surrounding the target word.

"Negative words" pertains to the use of words that trigger biased perceptions among readers. This paper introduces the concept of 'Words that Matter,' encompassing words that impart sentimental weight to a text. Price and Powers (1997) contend that journalists can significantly impact readers' perceptions, playing a pivotal role in shaping public opinion. Soroka and McAdams (2015) assert that the media's focus on negative news can adversely influence readers' attitudes. For example, if a news title contains a word with negative connotations, such as 'crisis,' 'recession,' or 'inflation', it can elicit negativity in readers' minds, even if the news itself is positive. Consequently, the influenced readers may exercise greater caution in their economic behaviors, and e.g., reduce their consumption. Our study develops a lexicon centered around the concept of 'Words that Matter'. Building upon the existing Kunsan National University lexicon (Kim, 2014), we construct a group of negative words using the keyword of 'crisis'. It is probably the most negative words of economy and can be a center of negative wordss to negative. By incorporating more relevant words and progressively adding biased terms, the newly added words might exhibit lower similarity to the central negative term. Consequently, the core negative term would possess stronger sentiment, acting as a focal point for negative words within the economic domain. The KNU sentiment lexicon consists of 9,826 negative words and 4,863 positive words (see Appendix B). The cosine similarity is utilized to determine contextual similarity with the word, 'crisis'.

---

[2] Token refers to a sequence of characters that represents a unit of meaning in a text, sort of word, phrase, punctuation mark, etc. Tokenization is the process of splitting a text to tokens; for example, the sentence "I am late !" is split into "I", "am", "late", "!".

[3] Accessible in GitHub repository: https://github.com/open-korean-text/open-korean-text

[4] Finding the best hyperparameters for a Word2Vec model typically involves experimentation and optimization based on the specific task and dataset. In this case, we manually selected the hyperparameters based on our understanding of the data and problem. The chosen hyperparameters include vector_size=100, window=10, min_count=10, workers=100, epochs=300, and sg=1. Generally, using more than 100 vectors and 300 epochs is recommended for improved performance. The average length of news titles in the dataset is 9.8 words, with a maximum of 36 words. A small window size might not capture the relationships between words within a sentence, while a large window size could link unrelated words. Therefore, we chose a window size of 10 to balance these considerations. See Appendix F.

We add 942 negative words to the KNU sentiment lexicon, after excluding 58 overlapped words in top 1,000 negative lexicons that showed high cosine-similarity with the keyword 'crisis'. As a result, the 'Words that Matter' lexicon contains 10,768 (=9826+942) negative and 4,863 positive words. <Figure 1> depicts the embedded words in a 2-dimension space. The embedded words, represented in a 100-dimensional space, were reduced to a two-dimensional space using Principal Component Analysis (PCA) and then visualized for intuitive understanding.

**<Figure 1> Visualization of Embedded Negative words**

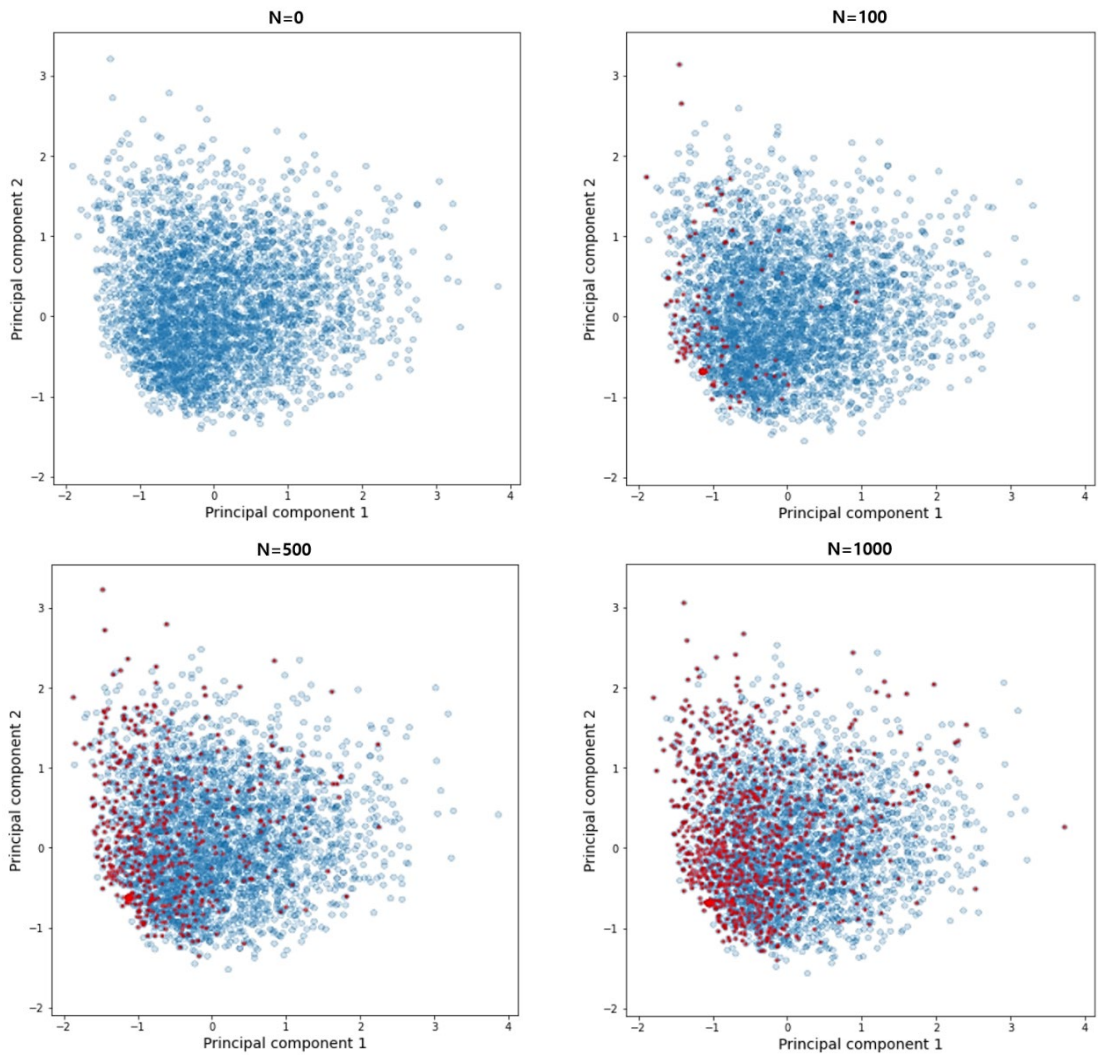

*Notes: 'N' indicates the number of negative lexicons, for instance, 'N=100' means a hundred of lexicons close proximity to keyword 'crisis'. The x-axis and y-axis are represented two principal component 1 and 2, Source: authors.

Equation (1) expresses our method to obtain news sentiment score from news titles. Each word is matched with new lexicons to determine whether it is positive or negative. Our sentiment score is computed to acquire a gap between the total number of positive and negative words. Each instance of a positive word is assigned a value of +1, while each instance of a negative word is assigned a value of -1. For instance, if a news title ($i$) contains two positive words ($P$) and four negative words ($N$), the sentiment score($SC_i$) can be -2.

$$SC_i = P_i - N_i \qquad (1)$$

$SC_i$ = sentiment score of $i^{th}$ news title
$P$ = number of positive words in the $i^{th}$ news title
$N$ = number of negative words in the $i^{th}$ news title

<Table 1> shows matching result in 45,723 news titles. As we add only negative words to dictionary, positive matching has fixed to 7,880 words, otherwise, negative matching has increased from 8,256 to 174,619 words. Thus, it is evident that the sum of words decreases, and the gap between the number of positive and negative words is significant.

**<Table 1> Lexicon Matching Result on Daily News Title**

| Threshold | Positive Matching ($P_i$) | - | Negative Matching ($N_i$) | = | Sum of words ($SC_i$) |
|---|---|---|---|---|---|
| N=0 | 7,880 words | - | 8,256 words | = | -376 |
| N=100 | 7,880 words | - | 60,802 words | = | -52,922 |
| N=500 | 7,880 words | - | 110,717 words | = | -102,837 |
| N=1,000 | 7,880 words | - | 174,619 words | = | -166,739 |

*Note: N=0 represents the baseline sentiment lexicon with no added negative lexicons. N=1000 refers to the inclusion of the top 1,000 negative words with high cosine similarity to the keyword 'crisis'.

This process increases the volatility of sentiment score by including the bag of negative words. As the threshold (N) is increased to 1,000, negative wordss gradually, sentiment becomes skewed to pessimism. However, when the number of negative words exceeds 30% of the total number of words, it becomes challenging to establish a semantic relationship between them. The quantitative optimization of this issue will be addressed in future studies, while the current study aims to examine the effect of negative words by comparing two separate cases: N=0 and N=1000. In the following section, we investigate the role of Heightened negativity and its ability to efficiently capture trends in the stock market. We expect that our sensitive detection of negative sentiment in news will move ahead of the stock market, which is a well-known leading indicator of the economy.

**Ordinary Least Squares regression: *i*-Step Forward Prediction**

To investigate the impact of news titles charged with heightened negativity on the KOSPI200 index (see Appendix C for summary statistics), this study employs the time series linear regression methods. KOSPI200 is an index of the 200 largest companies listed on the Korea Exchange (KRX). It is one of the major stock market indices in South Korea, alongside KOSPI and KOSDAQ. To enable comparison with the market index, we construct monthly time series indicators of news sentiment. This paper generates 'Sent0' and 'Sent1000' through the following process. We sum the sentiment scores of all news for each month. For Sent0, and Sent1000, respectively, we use lexicons based on N=0 and N=1,000 as discussed in the previous section. To extract the cyclical components, we apply the Hodrick-Prescott (HP) filter with lambda=14,400 (Hodrick and Prescott 1997; Ravn and Uhlig 2002)[5]. We then normalize the resulting series by minmax normalization $\frac{x-\min(x)}{\max(x)-\min(x)}$ so that the normalized values lie in the [0,1] interval. Lastly, a moving average is applied to reduce the noise and volatility of sentiment scores.

<Figure 2> visualizes the time series, providing a glimpse of the differences between Sent0 and Sent1000. Sent1000 demonstrates greater variability than Sent0 and frequently alters its trajectory prior to Sent0. This indicates that the sentiment indicator, derived from our novel lexicon supplemented with the top 1,000 negative terms, precedes its counterpart that relies on the conventionally employed KNU lexicon. In relation to KOSPI200, both Sent0 and Sent1000 fluctuate ahead of it. On balance, this observation suggests that Sent1000 could serve as a more effective predictor of KOSPI200 by detecting the shifts earlier than Sent0.

---

[5] For consistent analysis, we also apply the HP filter with lambda=14,000 to KOSPI200.

**<Figure 2> Visualization of Time Series Data**

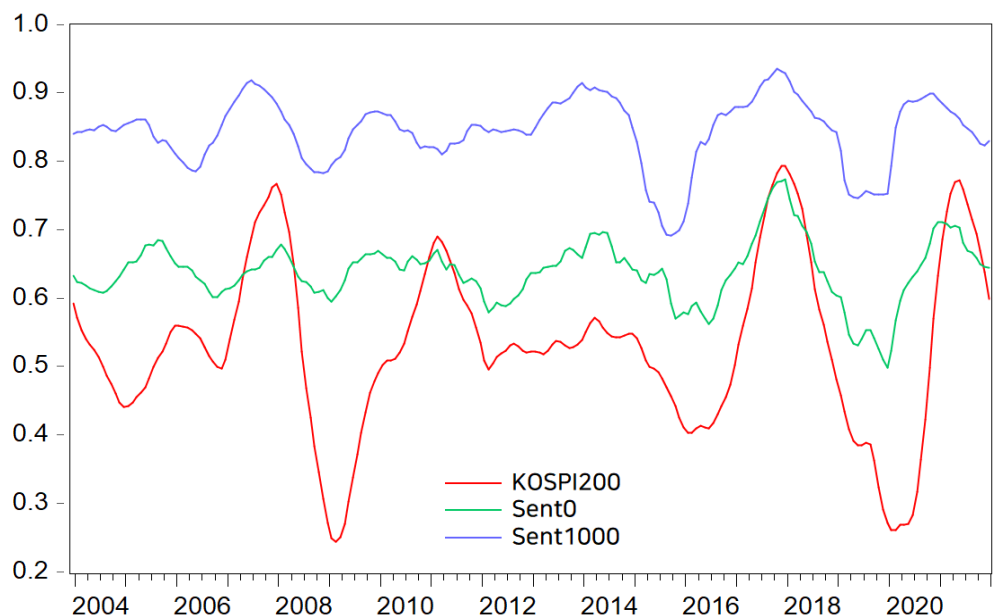

*Notes: Sent0 is the sentiment score based on the baseline KNU lexicon with no added words (N=0). Sent1000 is the sentiment score constructed using our expanded lexicon supplemented with top 1,000 negative words (N=1,000) with high cosine similarity to the keyword 'crisis'.

All filtered variables pass the unit root test and reject the null hypothesis of being a unit root process, indicating that the time series data are stationary. This implies that their statistical properties, such as mean and variance, remain constant over time, making them suitable for application of standard time series methods and modeling. We employ both time series ordinary least squares (OLS) and vector autoregression (VAR) techniques, as these methods are applicable to time series data that do not exhibit unit root characteristics. In the OLS model, we attempted to make *'i*-step forward predictions' to evaluate the extent to which time step contributes to forecasting performance of the stock market. Moreover, the use of VAR enables the examination of impulse response functions, which allows for the investigation of the effects of sentiment shocks on stock index.

The form of an OLS model with an AR(1) term can be expressed as Equation (2). In the model, independent variable (X) has tested for each time steps to check its impact on stock market index. Additionally, the AR(1) term can be expressed as a Wald decomposition, which enables the model to break down the total variance of the dependent variable. This can help control the serial correlation in the error terms by capturing the persistence or the effect of the past value of the dependent variable on its current value. If the coefficient of the AR(1) term($\beta$) approaches 1, it indicates that the long-run variance becomes dominant. Y is the dependent variable (KOSPI200), X is the independent variable (Sentiment Index), $Y_{-1}$ is the lagged dependent variable at one step before the present, $\alpha$ is intercept, and $\varepsilon$ is the error term. The model indicates how the relationship between KOSPI200 and Sentiment Index changes as the time steps of Sentiment Index (X) increase.

$$Y = \alpha + \beta\ Y_{-1} + \gamma\ X_{-i} + \varepsilon \qquad (2)$$

$Y$ : KOSPI200
$X$ : Sentiment Index (SENT0 or SENT1000)
$i$ : Time Step (0 to 8)

<Figure 3> illustrates the variations in coefficients for a lagged OLS model, examining the influence of different moving average windows. We investigated 1, 3, and 12-month windows and visually assessed the outcomes. We used centered moving average where the window is centered around the current time step (*t*). The window 1 moving average covers the time steps *t*-1 and *t*+1. Generally, SENT1000 demonstrates a positive correlation with KOSPI200, indicating that an increase in SENT1000 leads to a rise in KOSPI200. Conversely, SENT0 exhibits a negative association with KOSPI200. The coefficient of SENT1000 is consistently larger than that of SENT0 across most of lags. The positive relationship for SENT1000 persists for up to 6 months, whereas the positive association for SENT0 diminishes rapidly after 3 months. Given our focus on a negative sentiment, the positive correlation appears more plausible.

### <Figure 3> Variations of Coefficient ($\gamma$)

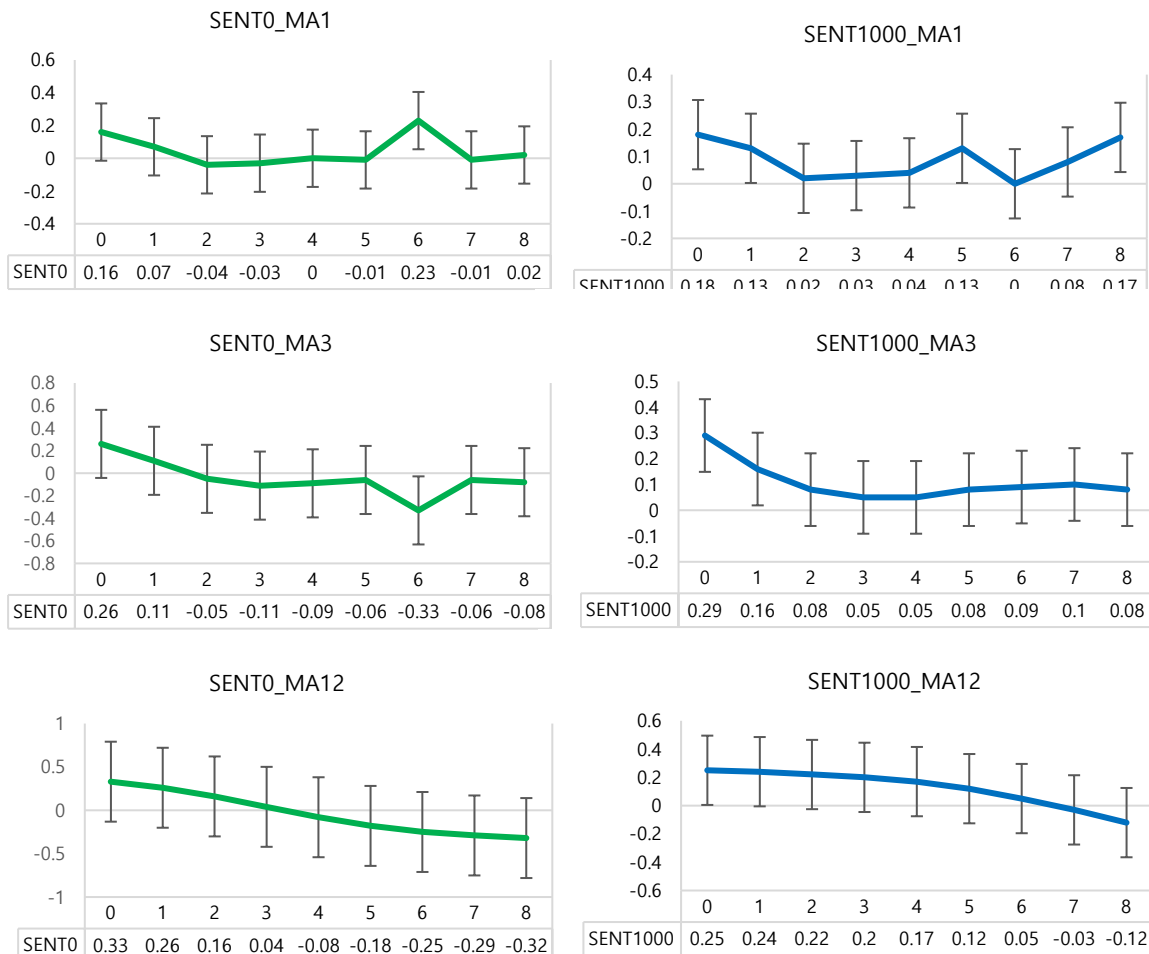

*Notes: The x-axis represents the number of time steps (lags, *i*); SENT0 represents the baseline sentiment index with no additional negativity, while SENT1000 indicates the intensified sentiment index with 1,000 negative lexicons; Each line on the graph includes two standard deviation bars; MA1 means the sentiment index transformed by moving average, while windows of 1,3,12 months were tesed; Please see appendix E dispalys detailed result of regression; Source by authors.

**Vector Auto Regression**

Vector Auto Regression (VAR) is a statistical method that is used for analyzing the dynamics between multiple interrelated time series. It helps in understanding the complex relationships among variables by accounting for their lagged interactions and simultaneous correlations. In the context of media impact on the stock market, VAR analysis recognizes the nuanced interplay between news dissemination and market behavior. However, the inherent nature of news sentiment introduces an endogenous challenge. News media can influence market movements and, at the same time, be influenced by prevailing market conditions, which can lead to content adjustments in response to economic indicators. In this scenario, the VAR model proves to be an apt approach for examining the influence of a news sentiment index on market dynamics. We have implemented a VAR model that incorporates the KOSPI200 ($Y$) and the News Sentiment Index ($X_0$ and $X_{1000}$) as depicted in equation (3), grounding our analysis in the assumption of media's multifaceted influence on the stock market.

$$M_t = \begin{bmatrix} Y_t \\ X_{0t} \\ X_{1000t} \end{bmatrix} = \alpha + \sum_{i=1}^{k} \beta_i M_{t-i} + \varepsilon_t \qquad (3)$$

where $M_t$ is a vector consisting of KOSPI200 ($Y_t$) and two kinds of sentiment index, $X_{0t}$ (=SENT0) and $X_{1000t}$ (=SENT1000). This paper uses three endogenous variables: $Y_t$ (KOSPI200), $X_0$ (SENT0), $X_{1000}$ (SENT1000), forming 3-dimensional vector. The model employed is a standard VAR with four lag intervals for endogenous variables. To select the appropriate number of lags, several lag selection criteria were used, including the Akaike Information Criterion (AIC), Bayesian Information Criterion (BIC), Hannan-Quinn Information Criterion (HQ), and Schwarz Information Criterion (SC). SC penalizes the number of parameters in the model more severely, , it can help us to select a model that does not consider too many time steps. This makes SC more effective in selecting parsimonious models with fewer parameters, which could help to avoid overfitting. In this study, based on the SC criterion, a model with four lag intervals was selected since we had only two independent variables, SENT0 and SENT1000. The results can be found in appendix H.

**Impulse Response Function**

As shown in studies by Sims (1980) and Bernanke and Gertler (1995), the IRF can also be used to investigate how the relationships between variables in the VAR model change over time. By estimating time varying IRFs, we can identify changes in the system's behavior and gain insights into the underlying causes of these changes. The IRF starts with one-time shock or innovation of the variables in the model. Usually, the result plotted graphically to visualize the dynamic response of variables to the shock.

<Figure 4> displays the response of KOSPI200 to shocks in Sent0 and Sent1000 under Cholesky decomposition assumption. One standard deviation shock in Sent0 has a minor negative impact on KOSPI200, which persists over time. Conversely, a one standard deviation shock in Sent1000 generates a significant positive effect that lasts until the 24-month periods. The IRF results corroborate the influence of Sent1000, which may be amplified by negative words in the analysis. The impact of negative lexicons demonstrates a greater magnitude and persistence compared to Sent0 and KOSPI200. This suggests that the influence of language with a negative or pessimistic tone has more substantial and long-lasting effects on the system than other factors under consideration. This emphasizes the importance of sentiment changes in shaping market dynamics and investor sentiment.

**<Figure 4> Response of KOSPI200 to Sentiment Index**

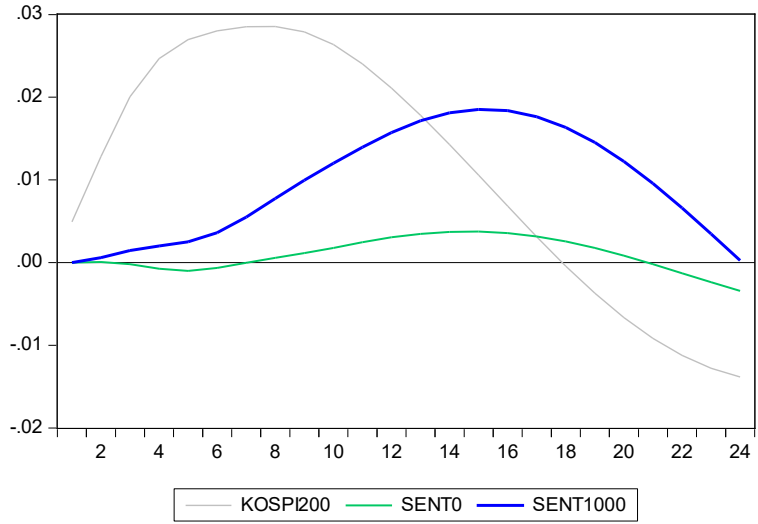

*Notes: x-axis represents the periods considered in impulse response function

Finally, <Figure 5> presents the results of the Granger Causality test (refer to Appendix D) that analyze the causal relationships between the variables in VAR model. For the first test, the dependent variable is KOSPI200, and the results show that Sent1000 is statistically significant at the 0.01 level of significance, while Sent0 is not statistically significant. This means that Sent1000 has a significant impact on KOSPI200, while Sent0 does not have a significant impact on KOSPI200. For the second test, the dependent variable is Sent0, and the results show that KOSPI200 and Sent1000 are both statistically significant at the 0.01 level of significance. This means that both KOSPI200 and Sent1000 have a significant impact on Sent0.

For the third test, the dependent variable is Sent1000, and the results show that neither KOSPI200 nor Sent0 is statistically significant. This means that KOSPI200 and Sent0 do not have a significant impact on Sent1000.

**<Figure 5> Illustration of Granger Causality Test**

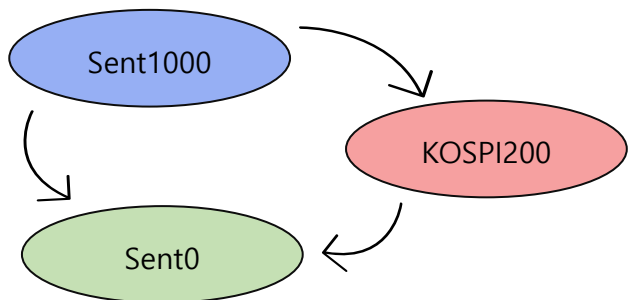

*Notes: x-axis represents the periods considered in impulse response function

Overall, these results suggest that Sent1000 has a significant impact on KOSPI200, while both KOSPI200 and Sent1000 have a significant impact on Sent0. However, KOSPI200 and Sent0 do not have a significant impact on Sent1000. These findings could be useful for understanding the relationship between sentiment and the stock market index and could be used for further analysis and forecasting.

**Conclusion**

In this study, we investigated the impact of negative words, specifically 'Words that Matter,' on sentiment analysis in a dataset of 45,723 daily economic news articles from South Korea. Employing Word2Vec, cosine similarity, and an expanded lexicon, we analyzed the influence of these words on the sentiment scores of news titles. Our findings reveal that incorporating negative words in the lexicon significantly amplifies sentiment scores' intensity, particularly the negativity of news titles. This suggests that the presence of emotionally charged words, such as those with high cosine similarity to the word 'Crisis,' considerably impacts the text's sentiment. It is crucial to understand the role of negative words in shaping readers' perceptions for accurate sentiment analysis. The influence of 'Words that Matter' underscores the importance of considering context and linguistic nuances when analyzing news content and its potential effects on public opinion and market dynamics.

In conclusion, this study examines the impact of news titles with heightened negativity on the KOSPI200 index using linear regression and sentiment analysis. The results indicate that the augmented sentiment lexicon (Sent1000), comprising the top 1,000 negative words with high cosine similarity to the keyword 'Crisis,' more effectively captures the impact of news sentiment on the stock market index than the original KNU sentiment lexicon (Sent0). The OLS model and Impulse Response Function (IRF) analyses disclose that Sent1000 has a more potent and persistent impact on KOSPI200 compared to Sent0. Furthermore, the Granger Causality test results demonstrate that Sent1000 displays predictive power for KOSPI200, with a significant impact over time, peaking at a 24-month horizon. Sent1000 also exerts considerable influence on both KOSPI200 and Sent0, driving changes in both variables. These findings emphasize the importance of understanding the role language plays in shaping market dynamics and investor sentiment, particularly the impact of negatively negative words on stock market indices. Future research could delve deeper into the underlying mechanisms through which news sentiment, especially those with heightened negativity, influences market behavior and investor decision-making.

**data availability statement**

The data that support the findings of this study are openly available in GitHub repository at https://github.com/wonseongkim/wtw.

**Appendix A.**

<Figure A.1> Word2Vec Architectures

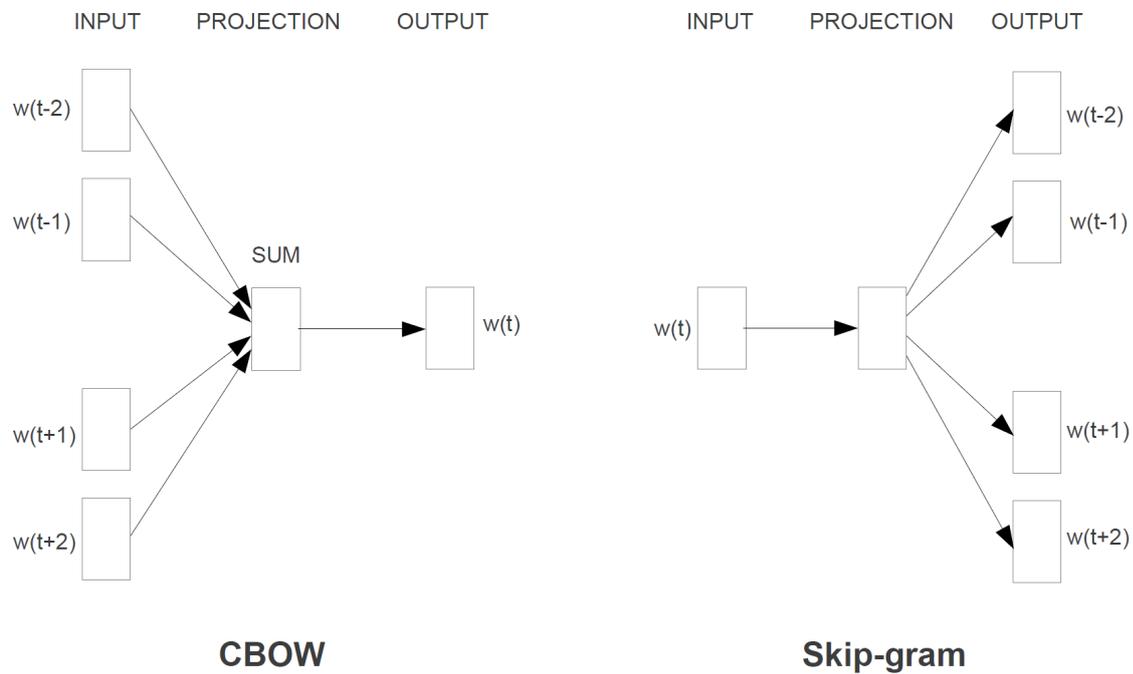

Mikolov et al. (2013) introduced the Word2Vec architectures Continuous Bag of Words (CBOW) and Skip-gram. <Figure A.1> depicts the structure of these models. The CBOW architecture predicts the target word based on its context, while the Skip-gram model predicts surrounding words given the target word.

<Table A.1> Comparison of architectures

| Model Architecture | Semantic-Syntactic Word Relationship test set | | MSR Word Relatedness Test Set [20] |
|---|---|---|---|
| | Semantic Accuracy [%] | Syntactic Accuracy [%] | |
| RNNLM | 9 | 36 | 35 |
| NNLM | 23 | 53 | 47 |
| CBOW | 24 | 64 | 61 |
| Skip-gram | 55 | 59 | 56 |

*RNNLM: Recurrent Neural Net Language Model / NNLM: New Log-linear Models (Mikolov et al., 2013)

The Skip-gram architecture performs slightly worse on syntactic tasks (59%<64%) than the CBOW model, and much better on semantic tasks (55%>24%) than all other models (Mikolov et al., 2013). The reason for choosing the Skip-gram architecture in this paper is its focus on finding semantic similarities between words rather than grammatical relationships.

**Appendix B.**

**<Figure B.1> Flow chart of 'KNU Sentiment Lexicon' construct algorithms**

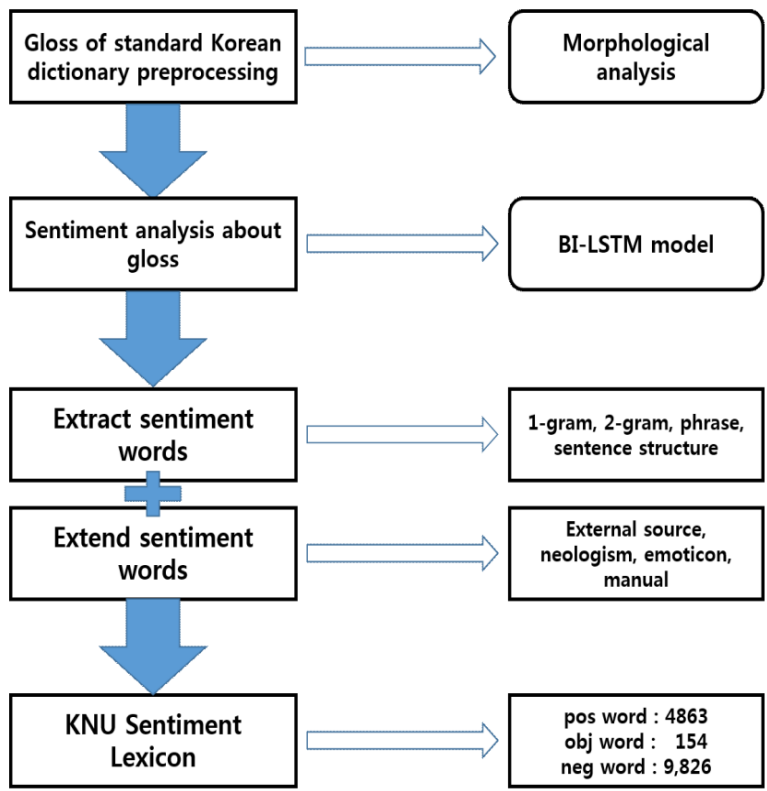

The KNU sentiment lexicon is a word-level NLP method that creates a Korean sentiment word dictionary. <Figure B.1> illustrates the process of constructing the KNU sentiment lexicon. The lexicon is derived from the 'Standard Korean Language Dictionary (SKLD),' published by the National Institute of the Korean Language (NIKL). All words in the SKLD are scraped and tokenized to match their corresponding glosses. It is assumed that if a gloss exhibits positive sentiment, the sentence must contain at least one positive word. Three annotators manually created a dataset for the glosses. If all annotators agreed on the same sentiment for a gloss (i.e., a sentence), the gloss was added to the dataset. Conversely, if there was disagreement, the gloss was excluded from the dataset

(https://github.com/park1200656/KnuSentiLex).

**<Figure B.2> General Structure of Bidirectional Recurrent Neural Network**

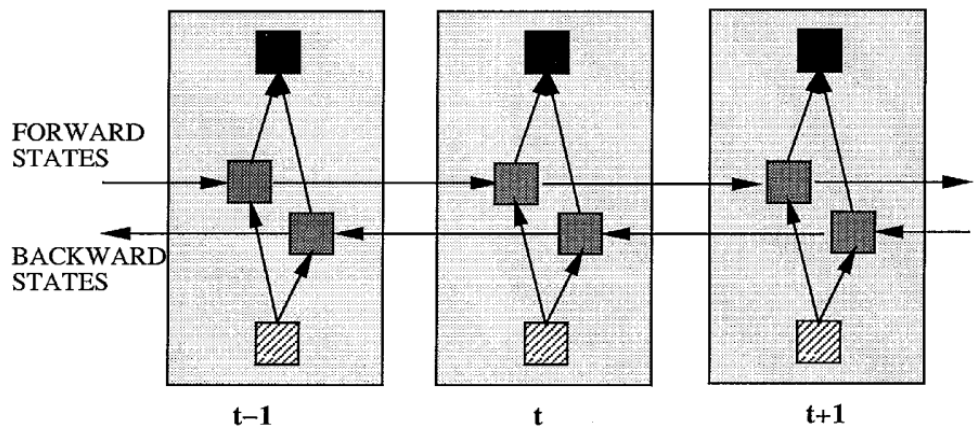

*source: Schuster and Paliwal (1997)

    The dataset was trained using a Bidirectional Long Short-Term Memory (Bi-LSTM) model. LSTM is a type of recurrent neural network designed for analyzing sequential data. Schuster and Paliwal (1997) introduced Bi-RNNs as powerful tools capable of considering both forward and backward states simultaneously <Figure A.2>. The hidden state in a Bi-LSTM captures both the left and right contexts. Based on the trained results, sentiment words (1-gram, 2-gram, phrases, and sentence structures) were extracted from the SKLD, resulting in the creation of the 'KNU sentiment dictionary'. The dictionary was further expanded with external sources, emoticons, and other elements. Ultimately, the dictionary comprises 4,863 positive words and 9,826 negative words.

**Appendix C.**

**<Table C.1> Descriptive statistics**

|  | Rawdata | | | Normalized Rawdata | | |
|---|---|---|---|---|---|---|
|  | KOSPI200 | Sent0 | Sent1000 | KOSPI200 | Sent0 | Sent1000 |
| Mean | 241.90 | -1.27 | -608.38 | 0.53 | 0.64 | 0.84 |
| Median | 248.41 | 0.00 | -527.00 | 0.53 | 0.64 | 0.85 |
| Maximum | 434.31 | 43.00 | -126.00 | 0.79 | 0.77 | 0.94 |
| Minimum | 96.50 | -100.00 | -2946.00 | 0.24 | 0.50 | 0.69 |
| Std. Dev. | 70.16 | 17.95 | 351.10 | 0.12 | 0.05 | 0.05 |
| Skewness | 0.25 | -1.74 | -2.60 | -0.05 | -0.02 | -0.81 |
| Kurtosis | 3.62 | 9.86 | 13.89 | 2.87 | 3.96 | 3.51 |
| Periods | 2003.12~2021.12 | | | 2003.06~2022.06 | | |
| Obs. | 217 | 217 | 217 | 229 | 229 | 229 |

*Sources: Authors' own computation. Data are taken from KOSPI200, KOSIS, Korea Press Foundation.

*Notes: Normalized Rawdata followed process: Rawdata – HP filtering (Cyclical factor extraction) – Minmax Normalizing – Centered moving average. With centered moving average, prieriods of data has been extended from 2003.12~2021.12 to 2003.06~2022.06, For instance, the calculation for a 3-period centered moving average (mavc) is as follows: mavc(x, period=3) = (x(-1) + x + x(1)) / 3. If the period is even, the window length is increased by one, and the two endpoints are weighted by 0.5 to achieve a balanced consideration. Additionally, any missing values (NA) are not propagated, and the dataset is automatically extended by adding raw data for six months to both sides of the original dataset.

Centered moving averages decrease the phase shift, or lag, associated with backward moving averages by factoring in data points before and after the current point. Additionally, centered moving averages reduce distortion compared to backward moving averages, which rely solely on past values. By considering both past and future values, centered moving averages offer a more balanced approach and lead to a more accurate depiction of the data.

**Appendix D.**

**<Table D.1> Result of VAR Granger Causality Tests**

Sample: 2003M06 2022M06
Included observations: 225

Dependent variable: **KOSPI200**

| Excluded | Chi-sq | df | Prob. |
|---|---|---|---|
| Sent0 | 1.281757 | 4 | 0.8645 |
| Sent1000 | 13.87944 | 4 | 0.0077 |
| All | 17.75075 | 8 | 0.0232 |

Dependent variable: **Sent0**

| Excluded | Chi-sq | df | Prob. |
|---|---|---|---|
| KOSPI200 | 23.68915 | 4 | 0.0001 |
| Sent1000 | 15.77883 | 4 | 0.0033 |
| All | 49.58315 | 8 | 0.0000 |

Dependent variable: **Sent1000**

| Excluded | Chi-sq | df | Prob. |
|---|---|---|---|
| KOSPI200 | 5.327539 | 4 | 0.2553 |
| Sent0 | 4.453703 | 4 | 0.3481 |
| All | 9.071419 | 8 | 0.3363 |

# Appendix E

### <Table E.1>

Linear Regression with Moving Averaged Sentiments

| Month | MA_12months | | MA_3months | | MA_1months | |
|---|---|---|---|---|---|---|
| Variable | SENT0 | SENT1000 | SENT0 | SENT1000 | SENT0 | SENT1000 |
| t=0 | 0.33 | 0.25 | 0.26 | 0.29 | 0.16 | 0.18 |
|  | (0.00) | (0.00) | (0.00) | (0.00) | (0.00) | (0.00) |
| t=1 | 0.26 | 0.24 | 0.11 | 0.16 | 0.07 | 0.13 |
|  | (0.00) | (0.00) | (0.03) | (0.00) | (0.14) | (0.01) |
| t=2 | 0.16 | 0.22 | -0.05 | 0.08 | -0.04 | 0.02 |
|  | (0.00) | (0.00) | (0.32) | (0.08) | (0.41) | (0.73) |
| t=3 | 0.04 | 0.2 | -0.11 | 0.05 | -0.03 | 0.03 |
|  | (0.42) | (0.00) | (0.02) | (0.30) | (0.45) | (0.52) |
| t=4 | -0.08 | 0.17 | -0.09 | 0.05 | -0.00 | 0.04 |
|  | (0.11) | (0.00) | (0.05) | (0.32) | (0.94) | (0.49) |
| t=5 | -0.18 | 0.12 | -0.06 | 0.08 | -0.01 | 0.13 |
|  | (0.00) | (0.00) | (0.20) | (0.10) | (0.89) | (0.01) |
| t=6 | -0.25 | 0.05 | -0.33 | 0.09 | 0.23 | 0.00 |
|  | (0.00) | (0.16) | (0.46) | (0.07) | (0.58) | (0.97) |
| t=7 | -0.29 | -0.03 | -0.06 | 0.10 | -0.01 | 0.08 |
|  | (0.00) | (0.48) | (0.21) | (0.04) | (0.85) | (0.15) |
| t=8 | -0.32 | -0.12 | -0.08 | 0.08 | 0.02 | 0.17 |
|  | (0.00) | (0.00) | (0.08) | (0.10) | (0.73) | (0.00) |

**Appendix F.**

**<Figure F.1> Distribution of News Titles by Length**

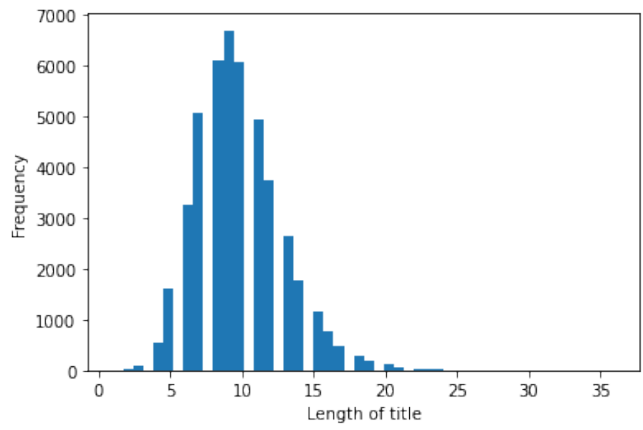

# Appendix G.

## <Table G.1> Vector Autoregression Estimates

Vector Autoregression Estimates
Sample (adjusted): 2003M10 2022M06
Included observations: 225 after adjustments
Standard errors in ( ) & t-statistics in [ ]

|  | KOSPI200 | SENT0 | SENT1000 |
|---|---|---|---|
| KOSPI200(-1) | 2.584318 | 0.184085 | 0.104550 |
|  | (0.06498) | (0.09390) | (0.07961) |
|  | [ 39.7719] | [ 1.96049] | [ 1.31321] |
| KOSPI200(-2) | -2.665277 | -0.173472 | -0.117791 |
|  | (0.16531) | (0.23888) | (0.20255) |
|  | [-16.1228] | [-0.72618] | [-0.58155] |
| KOSPI200(-3) | 1.446225 | -0.103748 | -0.039995 |
|  | (0.16591) | (0.23974) | (0.20328) |
|  | [ 8.71710] | [-0.43275] | [-0.19675] |
| KOSPI200(-4) | -0.394165 | 0.094350 | 0.061116 |
|  | (0.06495) | (0.09386) | (0.07958) |
|  | [-6.06870] | [ 1.00525] | [ 0.76798] |
| SENT0(-1) | -0.018578 | 1.651376 | -0.029695 |
|  | (0.04781) | (0.06909) | (0.05858) |
|  | [-0.38860] | [ 23.9031] | [-0.50694] |
| SENT0 (-2) | -0.015720 | -1.290902 | -0.011954 |
|  | (0.08876) | (0.12827) | (0.10876) |
|  | [-0.17710] | [-10.0642] | [-0.10991] |
| SENT0 (-3) | 0.056671 | 0.896509 | 0.103758 |
|  | (0.08770) | (0.12673) | (0.10745) |
|  | [ 0.64621] | [ 7.07429] | [ 0.96563] |
| SENT0 (-4) | -0.025054 | -0.337817 | -0.084249 |
|  | (0.04513) | (0.06521) | (0.05529) |
|  | [-0.55518] | [-5.18033] | [-1.52371] |
| SENT1000(-1) | 0.108271 | 0.209677 | 2.078945 |
|  | (0.05458) | (0.07887) | (0.06687) |
|  | [ 1.98378] | [ 2.65858] | [ 31.0887] |
| SENT1000 (-2) | -0.240282 | -0.224211 | -1.816180 |
|  | (0.11525) | (0.16655) | (0.14121) |
|  | [-2.08481] | [-1.34622] | [-12.8612] |
| SENT1000 (-3) | 0.217702 | -0.036747 | 1.087451 |
|  | (0.11408) | (0.16486) | (0.13978) |
|  | [ 1.90828] | [-0.22290] | [ 7.77980] |
| SENT1000 (-4) | -0.053458 | 0.098708 | -0.393212 |
|  | (0.05241) | (0.07574) | (0.06422) |
|  | [-1.01999] | [ 1.30331] | [-6.12330] |
| C | -0.010232 | 0.011072 | 0.046081 |
|  | (0.00818) | (0.01182) | (0.01002) |
|  | [-1.25076] | [ 0.93662] | [ 4.59736] |
| R-squared | 0.998544 | 0.976847 | 0.986291 |
| Adj. R-squared | 0.998462 | 0.975537 | 0.985515 |

**Appendix H.**

**<Table H.1> Model Selection Criteria for Optimal Lag Length**

| Lag | AIC | BIC | HQ |
|---|---|---|---|
| 0 | -8.947080 | -8.900353 | -8.928204 |
| 1 | -17.90195 | -17.71504 | -17.82645 |
| 2 | -21.23913 | -20.91205 | -21.10700 |
| 3 | -21.73690 | -21.26964 | -21.54815 |
| 4 | -22.22518 | **-21.61774*** | -21.97980 |
| 5 | -22.35182 | -21.60419 | -22.04981 |
| 6 | -22.49156 | -21.60376 | -22.13293 |
| 7 | -22.49881 | -21.47082 | -22.08355 |
| 8 | -22.63550 | -21.46734 | **-22.16361*** |
| 9 | -22.57197 | -21.26362 | -22.04345 |
| 10 | -22.60308 | -21.15455 | -22.01793 |
| 11 | -22.66053 | -21.07183 | -22.01876 |
| 12 | **-22.80771*** | -21.07882 | -22.10931 |

AIC: Akaike information criterion
BIC: Bayesian Information Criterion
HQ: Hannan-Quinn information criterion
* indicates lag order selected by the criterion